\pdfoutput=1

\documentclass[11pt]{article}

\usepackage[]{ACL2023}

\usepackage{times}
\usepackage{latexsym}
\usepackage{graphicx}
\usepackage{multirow}

\usepackage[T1]{fontenc}

\usepackage[utf8]{inputenc}

\usepackage{microtype}

\usepackage{inconsolata}
\usepackage{multirow}
\usepackage{graphicx}
\usepackage{enumitem}
\usepackage{listings}
\usepackage[ruled]{algorithm2e}
\usepackage{amsmath,amsthm,amsfonts,amssymb,bm}

\title{Data Uncertainty-Aware Learning for Multimodal Aspect-based Sentiment Analysis}

\author{Hao Yang, Zhenyu Zhang, Yanyan Zhao\thanks{* Corresponding author} \and Bing Qin \\
        Harbin Institute of Technology \\
        hyang@ir.hit.edu.cn, zyzhang@ir.hit.edu.cn, yyzhao@ir.hit.edu.cn, qinb@ir.hit.edu.cn}

\begin{document}
\maketitle
\begin{abstract}

As a fine-grained task, multimodal aspect-based sentiment analysis (MABSA) mainly focuses on identifying aspect-level sentiment information in the text-image pair. However, we observe that it is difficult to recognize the sentiment of aspects in low-quality samples, such as those with low-resolution images that tend to contain noise. And in the real world, the quality of data usually varies for different samples, such noise is called data uncertainty. But previous works for the MABSA task treat different quality samples with the same importance and ignored the influence of data uncertainty. In this paper, we propose a novel data uncertainty-aware multimodal aspect-based sentiment analysis approach, UA-MABSA, which weighted the loss of different samples by the data quality and difficulty. UA-MABSA adopts a novel quality assessment strategy that takes into account both the image quality and the aspect-based cross-modal relevance, thus enabling the model to pay more attention to high-quality and challenging samples. Extensive experiments show that our method achieves state-of-the-art (SOTA) performance on the Twitter-2015 dataset. Further analysis demonstrates the effectiveness of the quality assessment strategy.
\end{abstract}

\section{Introduction}

Fine-grained multimodal sentiment analysis aims to select fine-grained aspects from image-text pairs and determine their sentiment polarities. An integrated fine-grained multimodal sentiment analysis system can be formulated as four stages: 1) \textbf{data acquisition} aims to gather and select image-text pairs with sentiment tendencies from real-world social media for analysis purposes. 2) \textbf{aspect extraction} aims to identify the fine-grained aspects in the textual content \cite{sun2020riva,yu2020improving,wu2020multimodal,zhang2021multi,chen2022good}.
\begin{figure}[ht]
    \centering
    \includegraphics[width=7.5cm]{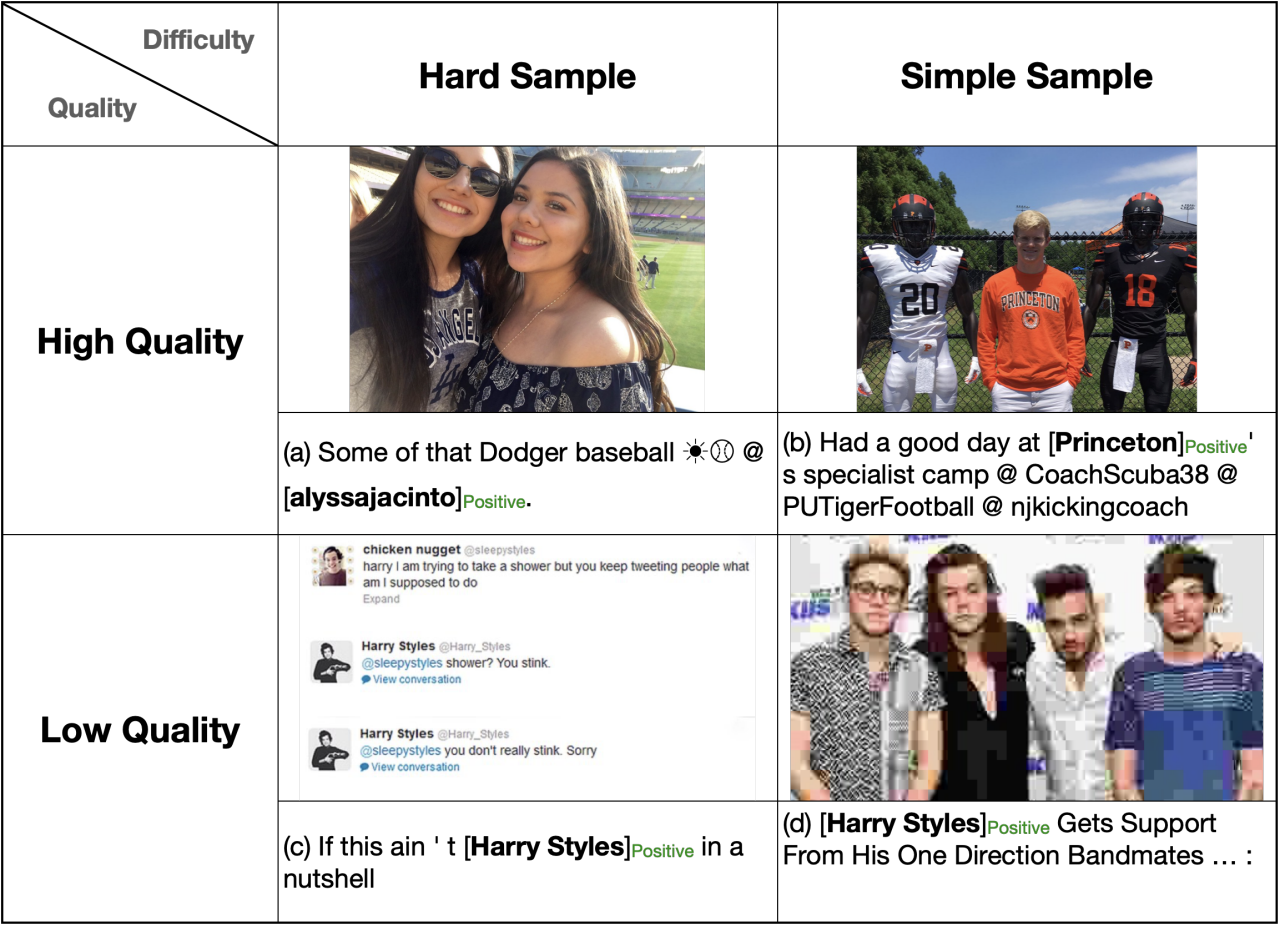}
    \caption{Examples of MABSA data with different quality and difficulty. High-quality and low-quality samples are distinguished by image recognizability and image-text correlation. Emotional recognition of hard samples requires cross-modal interaction information, while simple samples rely more on text content. }
    \label{case_first}
\end{figure}
3) \textbf{aspect-based sentiment analysis} aims to predict the sentiment polarity of the extracted aspects based on multimodal content \cite{xu2019multi,yu2019entity,yu2019adapting,khan2021exploiting,yang-etal-2022-face}. 4) \textbf{sentiment application} utilizes the predicted aspect-sentiment pairs in social media analysis, recommend systems and other relevant applications.

The collected image-text pairs data from uncontrollable open domains exhibit significant variability. Some data filtering methods have been applied in the first stage, such as filtering samples with long text or multiple images. However, existing methods ignore the data uncertainty issue caused by the variability, the sample with low visual recognizability misleads the model to learn more bias rather than the cross-modal interaction. As shown in Figure~\ref{case_first}, the visual content in example (c) is hard to recognize but be treated as same as the image in example (a), which introduce noise because the model is also unable to independently recognize sentiment from the text content. Therefore, we propose to introduce data uncertainty assessment to estimate the quality of multimodal fine-grained sentiment analysis data, and reduce the noise caused by data uncertainty. 



The existing fine-grained multimodal sentiment analysis work generally focuses on the samples like examples (a) and (b) in Figure~\ref{case_first}. But in fact, low-quality samples like example 1(c) and (d) widely exist and their visual features are ambiguous or absent. The core of the problem is how to identify and estimate the data quality in the MABSA task.

\citet{yang-etal-2022-face} proposed the FITE model, which utilizes facial information from images and achieves impressive results on the MABSA task. However, due to the sensitivity of facial expression recognition models to data noise, samples that can recognize facial expressions often have higher image quality. This method enhances the mining of visual sentiment information while relatively reducing the emphasis on low-quality noisy samples. Nevertheless, this process relies on external facial expression recognition models and does not provide a direct measure of data uncertainty.

In this paper, we propose incorporating two principles for evaluating the quality of multimodal sentiment samples: 1) considering the factor that affects the image quality, and 2) considering the relevance of fine-grained multimodal information. And we propose a data uncertainty-aware method UA-MABSA. The UA-MABSA comprehensively considers the assessment of sample quality based on visual ambiguity, correlation between image and text, and fine-grained information correlation between image and aspect, and combines them in a unified loss function. UA-MABSA adaptively changes the weight in loss function to assign different importance to different difficulties of samples, based on the sample quality.

We conduct extensive experiments on UA-MABSA in Twitter-2015 and Twitter-2017 datasets. Experiment results show that our model achieves state-of-the-art performance on multimodal aspect-based sentiment analysis task. The experiment results demonstrate that our proposed method effectively prevents the model from overfitting low-quality noisy samples. In addition, ablation experiments also demonstrate the effectiveness of different quality assessment strategies. To sum up, UA-MABSA improves previous work in three aspects:
\begin{itemize}
\item For the first time, we explored the data uncertainty and quality issues in fine-grained multimodal sentiment analysis tasks.
\item We verified a set of factors that affect multimodal data quality through extensive experiments. And we propose a sample quality assessment strategy that takes into account both the image quality and the aspect-based cross-modal relevance for multimodal aspect-based sentiment analysis.
\item We propose the UA-MABSA method, which adopts the proposed quality assessment strategy and effectively prevents model overfitting. UA-MABSA can be easily combined with previous methods and achieve new state-of-the-art results on the Twitter-2015 dataset. 
\end{itemize}

\section{Related Work}
\subsection{Fine-grained Multimodal Sentiment Analysis}
With the proliferation of online multimodal data, multimodal sentiment analysis has been extensively studied and tends to transition from coarse-grained to fine-grained. Coarse-grained multimodal sentiment analysis works \cite{yang2021multimodal,li2022clmlf} aims to detect the overall sentiment of each image-text pair. While the fine-grained multimodal sentiment analysis aims to select fine-grained aspects and detect the sentiment of the selected aspect based on the multimodal content. Previous studies normally cast fine-grained multimodal sentiment analysis as three sub-tasks, including Multimodal Aspect Term Extraction \cite{sun2020riva,yu2020improving,wu2020multimodal,zhang2021multi,chen2022good}, Multimodal Aspect Sentiment Classification \cite{xu2019multi,yu2019entity,yu2019adapting,khan2021exploiting,yang-etal-2022-face}, and Joint Multimodal Aspect-based Sentiment Analysis \cite{ju2021joint,yang2022cross,ling2022vision,yang2023few}.

Previous works have achieved impressive results for fine-grained multimodal sentiment analysis, but they ignored the impact of data uncertainty. Due to the multimodal data collected from the real-world open domain with weak cross-modal aligned supervision, the data uncertainty caused by low-quality samples will further make it more difficult to learn cross-modal alignment supervision. This has a negative impact on building a multimodal fine-grained sentiment analysis system for real-world multimodal data.

\begin{figure*}[ht]
    \centering
    \includegraphics[width=16cm]{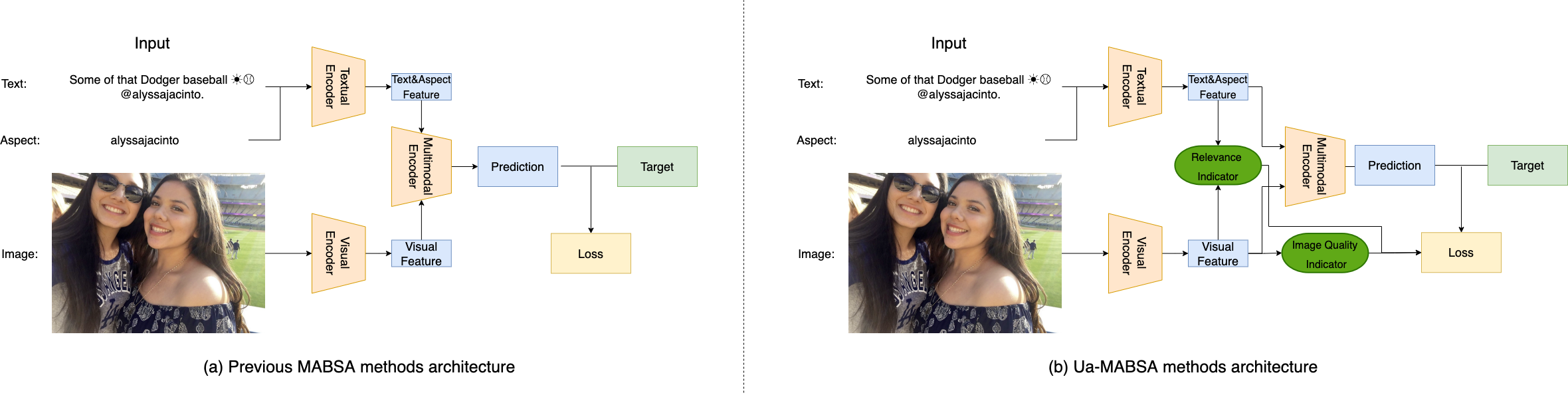}
    \caption{Previous MABSA methods structure vs UA-MABSA method structure: (a) Previous MABSA methods structure with uni-modal encoders and cross-modal encoder. (b) UA-MABSA method structure, the weight of loss is adjusted based on the image quality, image-text relevance and aspect-image relevance. }
    \label{fig-model}
\end{figure*}

\subsection{Data Uncertainty Learning}
Data uncertainty learning is used to capture the ``noise'' inherent in the data \cite{chang2020data}. As the noise widely exists in multimodal data, modeling data uncertainty is important for multimodal applications and has been used in the computer vision field, e.g., face recognition \cite{shi2019probabilistic,chang2020data,meng2021magface,kim2022adaface} and object detection  \cite{choi2019gaussian,kraus2019uncertainty}. 
Recently, some works \cite{blundell2015weight,gal2016dropout} have investigated how to estimate uncertainty in deep learning. By leveraging data uncertainty, models make significant advancements in robustness and interpretability. \citet{kendall2017uncertainties,kendall2018multi} utilize probabilistic models to capture data uncertainty and reduce the impact of noisy samples. Some works \cite{geng2021uncertainty,abdar2021review,loftus2022uncertainty,gour2022uncertainty,ji2023map} have also considered the data uncertainty in other multimodal tasks. Our method introduces data uncertainty into multimodal fine-grained sentiment analysis. With the help of uncertainty, the proposed model can automatically estimate the importance of different samples. Experimental results demonstrate that incorporating data uncertainty improves the performance of multimodal fine-grained sentiment analysis systems.

\section{Method}

\subsection{Task formulation}
The task of MABSA can be formulated as follows: given a set of multimodal samples $S = \lbrace X_{1}, X_{2},..., X_{|S|}\rbrace$, where $|S|$ is the number of samples. And for each sample, we are given an image $V \in \mathbb{R}^{3 \times H \times W} $ where $3$, $H$ and $W$ represent the number of channels, height and width of the image, and an $N$-word textual content $T = (w_{1},w_{2},...,w_{N})$ which contains an $M$-word sub-sequence as target aspect $A = (w_{1},w_{2},..,w_{M})$. Our goal is to learn a sentiment classifier to predict a sentiment label $y \in \lbrace Positive, Negative, Neutral\rbrace$ for each sample $X = (V, T, A)$.

\subsection{Overview}
Multimodal aspect-based sentiment analysis focuses on learning cross-modal fine-grained sentiment semantics from open-domain image-text pair data. Previous works proposed that this task faces two core challenges: 1) mining and utilizing visual emotional clues for the aspect. 2) cross-modal fine-grained alignment under weak supervision of image-text pair data. We propose that multimodal data uncertainty is also one of the core challenges of this task. For real-world applications, the noise in multimodal data is inevitable, which significantly affects the performance of MABSA models. 
Data uncertainty can directly lead to ineffective visual modality information mining and introduce a large amount of noise in the cross-modal fine-grained alignment learning stage. Therefore, it is necessary to introduce data uncertainty assessment in multimodal fine-grained sentiment analysis models.

As shown in Figure~\ref{fig-model}, based on the analysis of characteristics in multimodal fine-grained sentiment analysis data, we propose the data uncertainty-aware multimodal aspect-based sentiment analysis (UA-MABSA) model. The UA-MABSA model comprises three components: image-quality assessment, correlation assessment and backbone model. In the following subsections, we will provide detailed explanations of each component separately.

\begin{figure*}[ht]
    \centering
    \includegraphics[width=16cm]{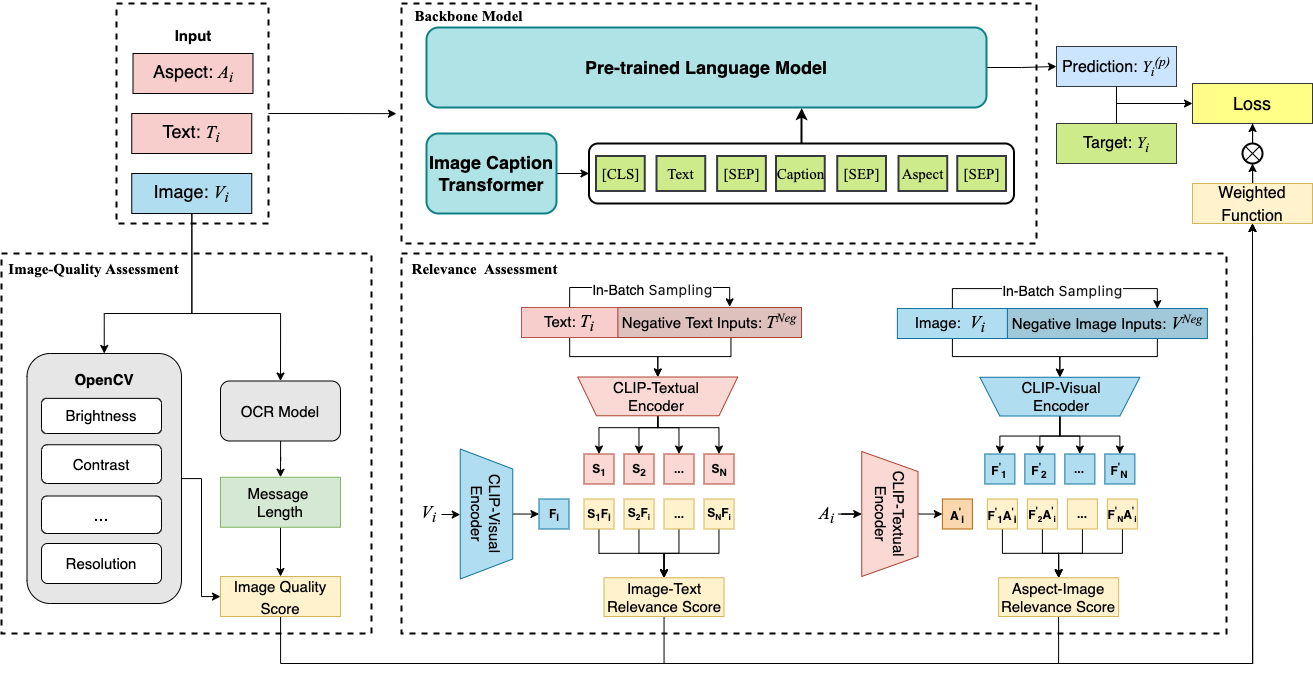}
    \caption{The overview of data uncertainty-aware multimodal aspect-based sentiment analysis(UA-MABSA) model architecture. }
    \label{fig-model}
\end{figure*}

\subsection{Image Quality Assessment}
Image quality is a combination of attributes that indicates how faithfully an image captures the original scene \cite{sheikh2006image}. Factors that affect the image quality include brightness, contrast, sharpness, color constancy, resolution, tone reproduction, etc. 
We comprehensively consider the brightness, contrast, sharpness, color constancy and resolution that affect image quality and used the tool provided by OpenCV \footnote{\url{https://github.com/opencv/opencv-python}} to calculate the score of the corresponding factors for each image. Taking image resolution as an example, we first set an image resolution threshold $t_{r}$ and calculate the resolution score $W_{i}^{r}$ of the image $V_{i}$ as follows:
\begin{gather}
W_{i}^{r} = \begin{cases} 
1 \quad q_{i} > t_{r} \\
\frac{q_{i}}{t_{r}}  \quad q_{i} \leq t_{r}
\end{cases}.\tag{1}
\end{gather}
where $q_{i}$ is the image resolution calculated by the OpenCV tool. When the image resolution $q_{i}$ is greater than the threshold $t_{r}$, the current image resolution score is set to 1. On the contrary, the image resolution score is calculated based on the proportion of the current image resolution to the threshold $t_{r}$.

In addition, we found that some images contain a large amount of textual information, which is difficult to convert to effective image features by the visual encoder(eg.ResNet \cite{he2016deep}) or convert to image captions by caption models. Therefore, the textual information in the image is treated as additional noise in existing MABSA models. We identified textual information in the images through OCR tools(PaddleOCR\footnote{\url{https://github.com/PaddlePaddle/PaddleOCR}}), but experiments have shown that the text information recognized by OCR does not significantly improve the effectiveness of MABSA task (see section 4 for details). Therefore, we only use the length of textual information in the image as a factor to measure the quality of the image. The formula for calculating the score of textual information in the image is as follows:
\begin{gather}
W_{i}^{text} = \begin{cases} 
1 \quad\quad\quad\quad L_{i} \leq t_{text}\\
1-\frac{L_{i}}{L_{max}}  \quad L_{i} > t_{text} 
\end{cases}.\tag{2}
\end{gather}
where $L_{i}$ represents the length of OCR result of image $V_{i}$, and $t_{text}$ is the artificially set threshold, and $L_{max}$ is the maximum length of OCR result of all images in the dataset.

Taking into account the above factors, we calculate a comprehensive score $W_{i}^{Image}$ for image quality by averaging the scores of each factor.
\subsection{Correlation Assessment}
Different from purely image-based datasets in the computer vision field, the interaction between text and image plays a crucial role in multimodal aspect-based sentiment analysis datasets. Considering the characteristics of multimodal aspect-based sentiment analysis data, we propose two correlation assessment methods: coarse-grained correlation assessment and fine-grained correlation assessment, based on the vision-and-language pretraining model CLIP \cite{radford2021learning}.

\textbf{Coarse-grained correlation assessment} primarily focuses on the correlation between image and text. Since the association between image and text varies, some samples may have limited additional information provided by the image. Therefore, the correlation between images and text is a crucial factor in assessing sample quality.
To calculate the correlation between image and text, for each image $V_{i}$, we employ an In-batch sampling strategy during the training process. We randomly sample negative text inputs $T^{Neg}$ from the same batch. Subsequently, we utilize the CLIP uni-modal encoder to obtain uni-modal features. 
\begin{gather}
F_{i} = Image\_Encoder(V_{i}) \tag{3}\\
S = Text\_Encoder([T_{i},T^{Neg}]) \tag{4}
\end{gather}
where $T_{i}$ is the text input, $F_{i}$ is the image feature, $S=\lbrace S_{1},S_{2},...,S_{N}\rbrace$ is the text features and $N$ is the in-batch sample size. 

And with L2-normalization, we calculate the cosine similarity score between the image and the text. We take the cosine similarity score as the coarse-grained correlation score $W_{i}^{IT}$.
\begin{gather}
W_{i}^{IT} = (L2(F_{i} \cdot W_{v}) \cdot L2(S_{1} \cdot W_{s})^{T})*e^{t} \tag{5}
\end{gather}
where $L2$ represents the L2-normalization function, $W_{v}$ and $W_{s}$ are learnable weights, and $t$ is the temperature scaling in the CLIP model.

\textbf{Fine-grained correlation assessment} focuses on the correlation between aspect and image. Since each sample's textual content may contain multiple aspects, but not all aspects necessarily appear in the image, the importance of additional visual information varies for different aspects. Previous methods assigned equal weights to different aspects within the same image-text pair, which to some extent induced the model to learn irrelevant biases instead of cross-modal fine-grained alignment. And similar to the coarse-grained correlation assessment, we utilize the CLIP model to calculate the fine-grained correlation score $W_{i}^{AI}$.
\begin{gather}
A^{'}_{i} = Text\_Encoder(A_{i}) \tag{6}\\
F^{'} = Image\_Encoder([V_{i},V^{Neg}]) \tag{7}\\
W_{i}^{AI} = (L2(A^{'}_{i} \cdot W^{'}_{a}) \cdot L2(F^{'}_{1} \cdot W^{'}_{v})^{T})*e^{t} \tag{8}
\end{gather}
where $A_{i}$ is the aspect input, $A^{'}_{i}$ is the aspect feature, $V^{Neg}$ is negative image inputs, $F^{'}=\lbrace F^{'}_{1},F^{'}_{2},...,F^{'}_{N}\rbrace$ is the image features and $N$ is the in-batch sample size, $W^{'}_{a}$ and $W^{'}_{v}$ are learnable weights, and $t$ is the temperature scaling in the CLIP model.

\subsection{Backbone Model}
The backbone model in UA-MABSA is inspired by the design of CapBERT \cite{khan2021exploiting} and FITE \cite{yang-etal-2022-face}, we adopt Caption Transformer to convert the image into a caption to overcome the semantic gap between different modalities. To achieve sufficient interaction in the text modal, we concatenate the image caption with text and aspect to form a new sentence. And we feed the new sentence to a pre-trained language model and fine-tune the model to obtain the pooler outputs $Y_{i}^{(p)}$ for aspect-based sentiment classification. We use the standard cross-entropy loss $\mathcal{L}$ to optimize all the parameters. And during the computation of the loss, we introduce weights based on image quality scores and correlation scores to obtain uncertainty-aware loss $\mathcal{L}^{'}$ and make the model more sensitive to high-quality and challenging samples.
\begin{gather}
\mathcal{L}^{'} = avg(W_{i}^{Image}+W_{i}^{IT}+W_{i}^{AI})*\mathcal{L} \tag{9}
\end{gather}
where $avg$ represent the average function.

\section{Experiment}

\begin{table*}[]
\scalebox{0.8}{
\begin{tabular}{lllllllllllll}
\hline
\multicolumn{1}{c}{} & \multicolumn{6}{c}{Twitter-2015}                               & \multicolumn{6}{c}{Twitter-2017}                               \\ \cline{2-13} 
Split                & \#POS & \#Neutral & \#NEG & Total & \#Aspects & \#Len & \#POS & \#Neutral & \#NEG & Total & \#Aspects & \#Len \\ \hline
Train                & 928      & 1,883    & 368      & 3,179  & 1.34         & 16.72  & 1,508     & 416     & 1,638     & 3,562  & 1.41         & 16.21  \\
Valid.           & 303      & 679     & 149      & 1,122  & 1.33         & 16.74  & 515      & 144     & 517      & 1,176  & 1.43         & 16.37  \\
Test                 & 317      & 607     & 113      & 1,037  & 1.35         & 17.05  & 493      & 168     & 573      & 1,234  & 1.45         & 16.38  \\ \hline
\end{tabular}
}
\caption{Statistics of two benchmark datasets for multimodal aspect-based sentiment analysis task.}
\label{dataset}
\end{table*}

\subsection{Experiment Setup}
We trained our model and measured its performance on the Twitter-2015 and Twitter-2017 datasets. These two datasets consist of multimodal tweets that are annotated the mentioned aspect in text content and the sentiment polarity of each aspect. Each multimodal tweet is composed of an image and a text that contains the target aspect. The detailed statistics of the two datasets are shown in Table\ref{dataset}. In addition, we set the model learning rate as 5e-5, the pre-trained model attention head as 12, the dropout rate as 0.1, the batch size as 16 and the fine-tuning epochs as 8, and the maximum text length is 256. We report the average results of 5 independent training runs for all our models. And all the models are implemented based on PyTorch with two NVIDIA TeslaV100 GPUs.

\subsection{Compared Baselines}
In this section, we compared the UA-MABSA model with the following models and reported the accuracy and Macro-F1 score in Table\ref{main_result}. 

We compare the method in the image-only setting: the Res-Target model which directly uses the visual feature of the input image from ResNet \cite{he2016deep}. As well as the text-only models: (1) LSTM. (2) MGAM, a multi-grained attention network\cite{fan2018multi} which fuses the target and text in multi-level. (3) BERT, the representative pre-trained language model~\cite{devlin2018bert}, which has strong text representation ability and can learn alignment between two arbitrary inputs. Moreover, the multimodal compared baselines include: (1) MIMN, The Multi-Interactive Memory Network ~\cite{xu2019multi} learn the interactive influences in cross-modality and self-modality. (2) ESAFN, an entity-sensitive attention and fusion network \cite{yu2019entity}. (3) VilBERT, a pre-trained Vision-Language model \cite{lu2019vilbert}, the target aspect is concatenated to the input text. (4) TomBERT,  the TomBERT~\cite{yu2019adapting} models the inter-modal interactions between visual and textual representations and adopts a Target-Image (TI) matching layer to obtain a target-sensitive visual. (5) CapBERT, a BERT-based method \cite{khan2021exploiting} which translates the image to caption and fuses the caption with input text-aspect pair through the auxiliary sentence. (6) CapBERT-DE, which replaces BERT to BERTweet \cite{bertweet} in CapBERT. (7) VLP-MABSA \cite{ling2022vision}, which is a task-specific pre-training vision-language model for MABSA. (8) KEF \cite{zhao-etal-2022-learning-adjective} adopts adjective-noun pairs extracted from the image to enhance the visual attention capability. (9) FITE \cite{yang-etal-2022-face} capture visual sentiment cues through facial expressions and achieve impressive performance.

\begin{table}[]
\scalebox{0.82}{
\begin{tabular}{lcccc}
\hline
\multirow{2}{*}{Method} & \multicolumn{2}{c}{Twitter-2015} & \multicolumn{2}{c}{Twitter-2017} \\ \cline{2-5} 
                        & Acc           & Macro-F1         & Acc           & Macro-F1         \\ \hline
                         & \multicolumn{4}{c}{Image Only}                                    \\ \cline{2-5} 
Res-Target & 59.88          & 46.48          & 58.59          & 53.98          \\ \hline
                        & \multicolumn{4}{c}{Text Only}                                       \\ \cline{2-5} 
LSTM                    & 70.30         & 63.43            & 61.67         & 57.97            \\
MGAN                    & 71.17         & 64.21            & 64.75         & 61.46            \\
BERT$^{*}$                   & 74.25         & 70.04            & 68.88         & 66.12            \\ \hline
                        & \multicolumn{4}{c}{Text and Image}                                  \\  \cline{2-5}
MIMN                    & 71.84         & 65.69            & 65.88         & 62.99            \\
ESAFN                   & 73.38         & 67.37            & 67.83         & 64.22            \\
VilBERT                 & 73.69         & 69.53            & 67.86         & 64.93            \\
TomBERT$^{*}$                 & 77.15         & 71.15            & 70.34         & 68.03            \\
CapBERT$^{*}$                 & 78.01         & 73.25            & 69.77         & 68.42            \\
FITE$^{*}$                    & 78.49         & 73.90            & 70.90         & 68.70            \\
KEF$^{*}$                   & 78.68         & 73.75            & 72.12         & 69.96            \\ 
VLP-MABSA  & 78.60          & 73.80          & \textbf{73.80}          & \textbf{71.80}          \\ \cline{2-5}
UA-TomBERT$^{*}$                   & 78.49         & 73.30            & 71.15         & 69.24            \\
UA-MABSA$^{*}$               & \textbf{78.88}         & \textbf{74.49}            & 71.85         & 70.16            \\ \hline
\end{tabular}
}
\caption{Experiment results for multimodal aspect-based sentiment analysis. $*$ denotes the results are from BERT-based models. }
\label{main_result}
\end{table}

\subsection{Main Results and Analysis}
We compare our methods with the above baseline models, Table~\ref{main_result} summarizes the main results for the Twitter-2015 and Twitter-2017 datasets.  Accuracy (Acc) and Macro-F1 score are used for evaluation. For a fair comparison, we do not give the result of BERTweet-based models \cite{nguyen2020bertweet} which outperform BERT-based models by using additional domain-specific Tweet data for pre-training. In addition, in order to test the transferability of our proposed UA-MABSA method, we used TomBERT as the backbone model to validate the effectiveness of data uncertainty in cross-modal attention-based models. The best scores on each metric are marked in bold.

As shown in the Table~\ref{main_result}, we can make a couple of observations:
(1) our proposed methods outperform the baseline multimodal aspect-based sentiment analysis models on most benchmarks and achieve new SOTA results of the macro-F1 score on the Twitter-2015 dataset. This demonstrates the effectiveness of the proposed data uncertainty-aware multimodal aspect-based sentiment analysis method.
(2) The UA-TomBERT model, which introduces our multimodal data uncertainty assessment, shows improvement over typical approaches and outperforms the TomBERT model on the accuracy and macro-F1 score by 1.34\% and 2.15\% on the Twitter-2015 dataset, 0.81\% and 1.21\% on Twitter-2017 dataset. Similarly, UA-MABSA also performs better than CapBERT and FITE models. The experimental results demonstrate the superiority of the proposed data uncertainty-aware method on flexibility and the robustness to overfitting.
(3) Our methods also show competitive performances compared with the VLP-MABSA model. In contrast, UA-MABSA is simpler and has good potential for further enhancing performance.
(4) And compared with the baseline model, the improvement of the UA-MABSA model on the Twitter-2015 dataset is more significant than the improvement on the Twitter-2017 dataset. We conjecture this is because data uncertainty-aware learning is more suitable for the dataset with low proportion of low-quality data, but the proportion of low-quality data in the Twitter-2017 dataset is high.

\begin{table}[]
\scalebox{0.75}{
\begin{tabular}{lcccc}
\hline
\multirow{2}{*}{Method}    & \multicolumn{2}{c}{Twitter-2015} & \multicolumn{2}{c}{Twitter-2017} \\ \cline{2-5} 
                           & Acc             & Macro-F1       & Acc             & Macro-F1       \\ \hline
UA-MABSA                   & \textbf{78.88}  & \textbf{74.49} & \textbf{71.85}  & \textbf{70.16} \\
w/o Image Quality          & 77.63           & 73.52          & 69.88           & 68.89          \\
w/o I-T Relevance   & 77.82           & 73.64          & 69.57           & 68.44          \\
w/o A-I Relevance & 77.12           & 73.28          & 69.67           & 68.70          \\ 
+ OCR Result & 76.56           & 70.24          & 68.42           & 67.36          \\ \hline
UA-TomBERT                   & 78.49         & 73.30            & 71.15         & 69.24            \\
w/o Image Quality          & 77.43           & 72.40          & 66.93           & 66.96          \\
w/o I-T Relevance   & 77.24           & 71.39          & 70.50           & 67.76          \\
w/o A-I Relevance & 76.37           & 71.49          & 70.42           & 68.00          \\
+ OCR Result & 75.36           & 69.49          & 67.32           & 64.92          \\\hline
\end{tabular}
}
\caption{Ablation study of the proposed UA-MABSA model and UA-TomBERT model.}
\label{abla_result}
\end{table}

\begin{figure}[ht]
    \centering
    \includegraphics[width=7.5cm]{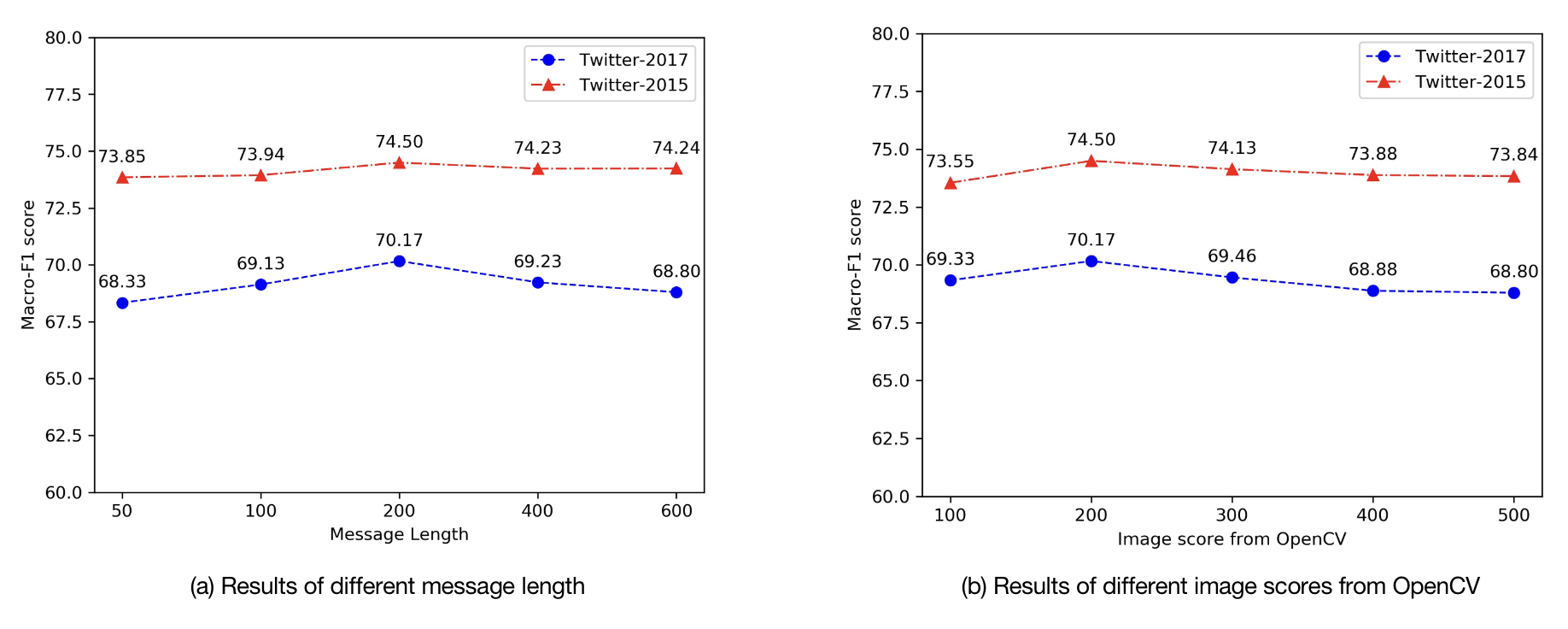}
    \caption{The performance of UA-MABSA with different thresholds of OCR message length and OpenCV image score for multimodal aspect-based sentiment analysis. }
    \label{thresholds}
\end{figure}

\begin{figure}[ht]
    \centering
    \includegraphics[width=8cm]{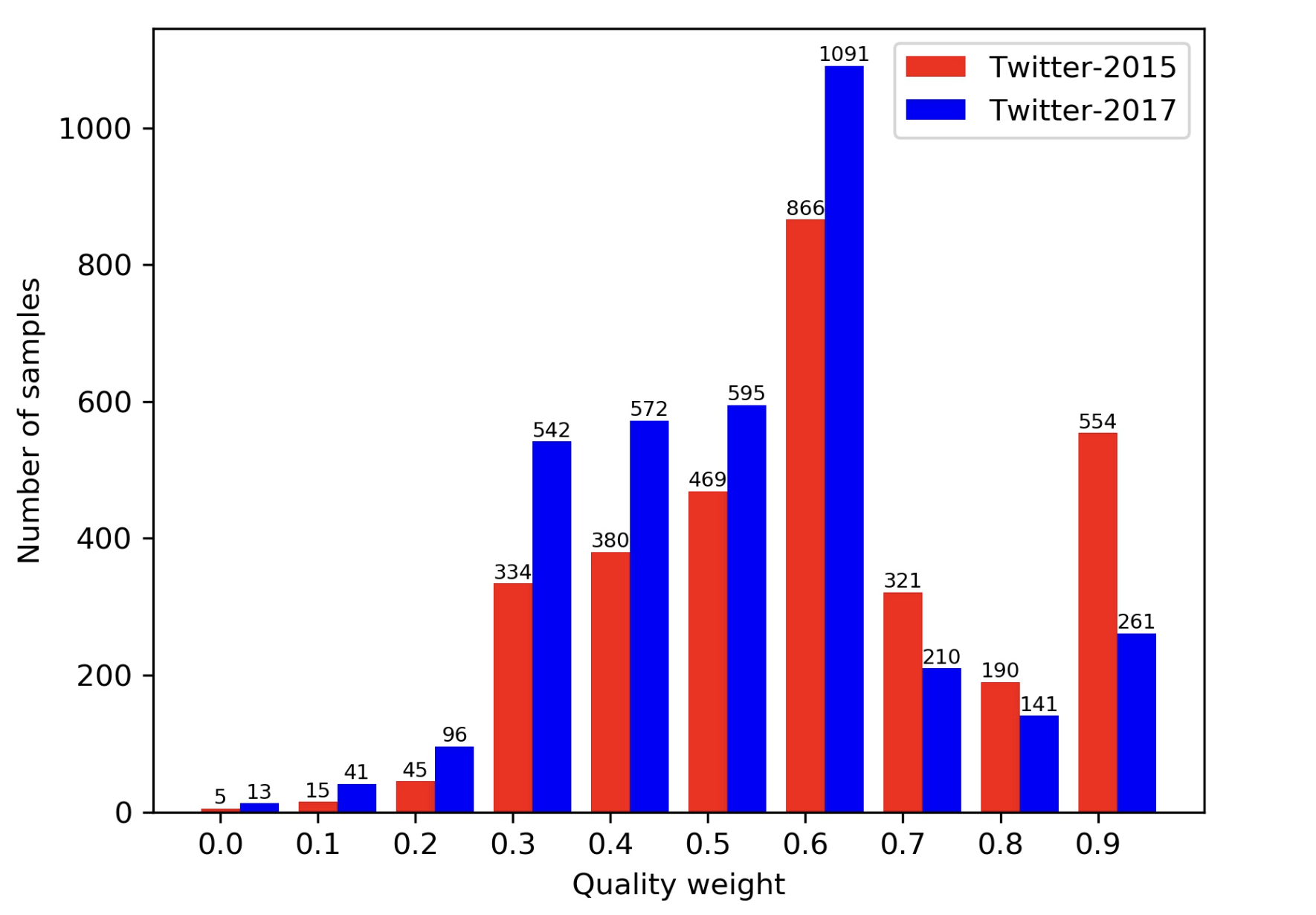}
    \caption{Comparison of quality weights between the Twitter 2015 dataset and the Twitter 2017 dataset. }
    \label{quality}
\end{figure}

\begin{figure*}[ht]
    \centering
    \includegraphics[width=15cm]{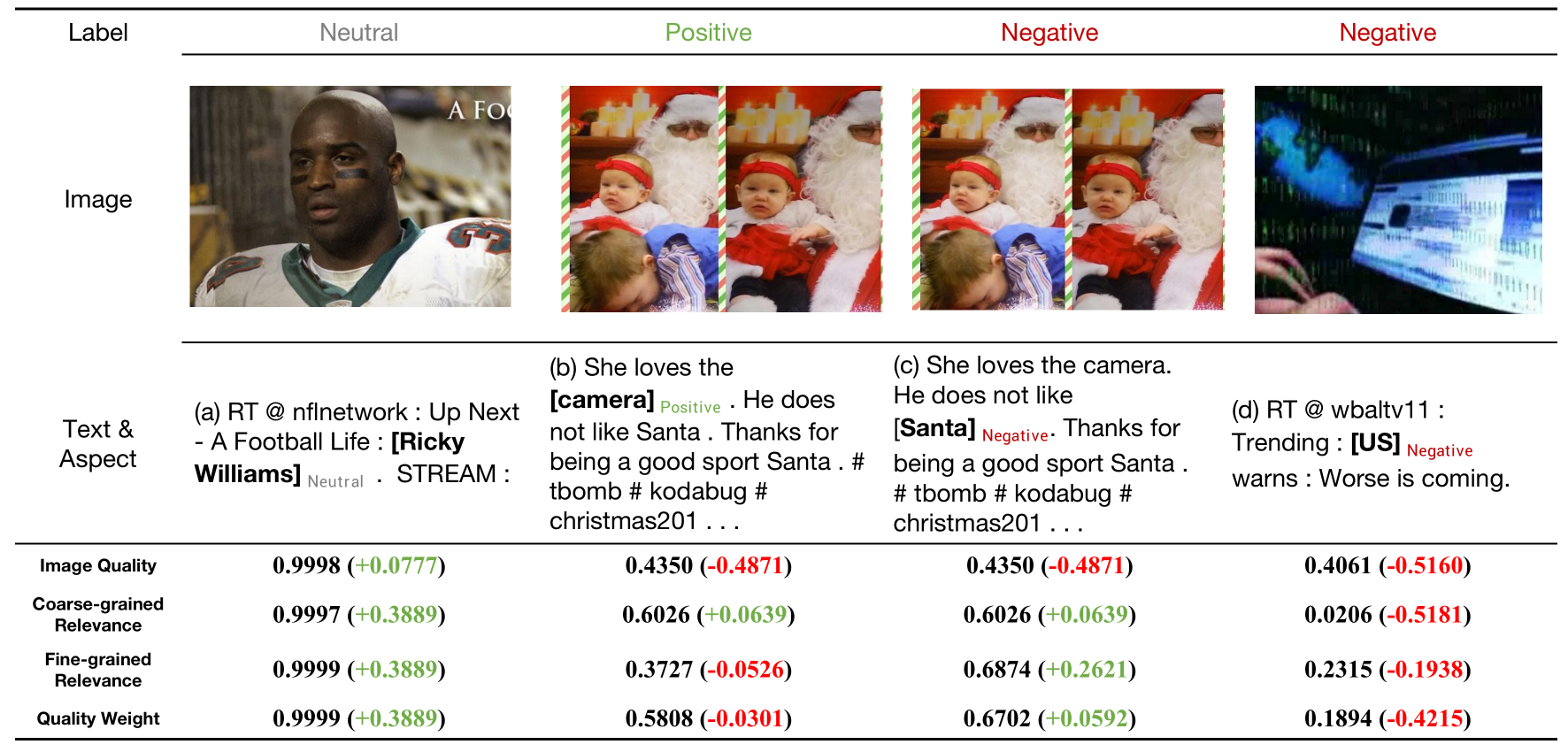}
    \caption{Case analysis of data uncertainty on image quality, coarse-grained and fine-grained relevance score, and sample quality weight. The red and green numbers represent the difference between the score and the average score of the datasets. }
    \label{case-study}
\end{figure*}

\subsection{Uncertainty Assessment Analysis}
In this part, we analyze the proposed UA-MABSA to gain more insights about data uncertainty learning.

Firstly, to explore the impact of each proposed sample quality assessment strategy of multimodal aspect-based sentiment analysis, we perform an ablation study using the UA-MABSA model on Twitter-2015 and Twitter-2017 datasets. The results are shown in Table~\ref{abla_result}. We calculated the new sample weights separately, removing the image quality score (w/o image quality), image-text relevance score (w/o I-T relevance), or aspect-image relevance score(w/o A-I relevance), to test the effectiveness of each strategy. In addition, to verify the effectiveness of each strategy under different backbone models, we tested both the UA-MABSA model and the UA-TomBERT model. After removing any quality assessment strategy, the effectiveness of both models decreased, which proves the effectiveness of our proposed data uncertainty-aware method. And for the Twitter 2015 dataset, removing the A-I relevance score resulted in the greatest decrease, with UA-MABSA decreasing by 1.21\% in the Macro-F1 score and UA-TomBERT decreasing by 1.81\%. While for the Twitter 2017 dataset, the image quality score plays a more important role. This also demonstrates the phenomenon of data uncertainty for different datasets.

Secondly, in the image quality assessment module, we manually set some thresholds, one for the text length recognized from the image using the OCR tool, and the other for the score of the Opencv tool that recognizes image resolution, brightness, etc.  In order to choose a reasonable threshold, we selected UA-MABSA models under different thresholds for analysis. As shown in Figure~\ref{thresholds}, We found that setting the threshold of resolution score and the threshold of OCR message length to 200 works best. The remaining thresholds can refer to the appendix. In addition, we also attempted to add the OCR results as an additional sentence to the model. But as shown in Table~\ref{abla_result}, the effect decreased significantly. We believe that the reason is that the impact of a large amount of noise information introduced by incorrect recognition in OCR exceeds the gain of supplementing additional semantic information.

Thirdly, to visualize the data uncertainty in Twitter 2015 and Twitter 2017 datasets, we presented the proportions of data with different quality weights. As shown in Figure~\ref{quality}, most of the data quality was within the range of 0.3 to 0.7, and the weight distribution of the two datasets was similar. Our proposed method can effectively capture uncertainty in the datasets and enable the model to pay varying degrees of attention to samples of different qualities, improving the robustness of the model to low-quality samples.

\subsection{Case Study}
In order to better understand the advantages of introducing data uncertainty in multimodal fine-grained sentiment analysis, we randomly selected some samples from the Twitter dataset for the case study.  Figure~\ref{case-study} shows the image quality scores, coarse-grained and fine-grained correlation scores, and comprehensive quality weight scores of different quality images calculated by the UA-MABSA method. Firstly, by comparing the image in the samples, we can find that the image quality assessment strategy can effectively distinguish between images of different qualities, such as high-quality images like the sample (a) and low-quality images like the sample (b), (c) and (d). Secondly, by comparing the sample (b) and (c) in Figure~\ref{case-study}, it can be found that for the same image-text pair with different aspect, the correlation and importance of the images is also varied. In sample (c), the fine-grained correlation is 0.6874, which is higher than the average value of 0.2621, while in sample (b), the fine-grained correlation is 0.3727, which is lower than the average value of -0.0526. UA-MABSA can capture this difference through fine-grained correlation and provide different weight scores. Thirdly, sample (d) shows a significantly low-quality example where the images in the sample are blurry and have a low correlation with the aspect, making it difficult for the model to obtain effective information from the image. Therefore, after considering three assessment strategies comprehensively, it is treated as a low-quality sample.

\section{Conclusion}
In this paper, we first identify the challenge of data uncertainty in the multimodal fine-grained sentiment analysis framework. And we define a quality assessment strategy for multimodal fine-grained sentiment analysis data to alleviate this issue, which takes into account the image's inherent quality and the coarse-grained and fine-grained relevance of the multimodal data. The proposed quality assessment strategy can also provide a reference for other multimodal tasks. And we propose the UA-MABSA method, which leads the model to pay more attention to high-quality and challenging samples and effectively prevents model overfitting. Extensive experiments demonstrate that our proposed method achieves competitive performance on the Twitter-2015 and Twitter-2017 datasets.

\section*{Limitations}
Although our method has shown superior performance, there are still a few limitations that could be improved in future work.
The major limitation is that our method has not yet been experimentally tested on aspect extraction and joint multimodal aspect-based sentiment analysis tasks. One of the main reasons is that the multimodal aspect-based sentiment analysis task is a representative and challenging sub-task in multimodal fine-grained sentiment analysis framework, and the performance on this sub-task can demonstrate the effectiveness of our data uncertainty-aware method. We will explore the data uncertainty in the aspect extraction task, joint multimodal aspect-based sentiment analysis task and few-shot multimodal aspect-based sentiment analysis task in future work. 
Another limitation is that in our UA-MABSA method, the assessment of sample quality still requires manual threshold judgment. And due to the small size of the dataset, it is difficult to learn image and text alignment supervision, we have to rely on the vision-and-language pre-trained model CLIP to provide additional supervision. In the future, we will explore more adaptive quality assessment methods.
\section*{Ethics Statement}
Our work complies with Twitter's data policy, and all the codes and datasets used in our work comply with the ethics policy. 
\bibliography{custom}
\bibliographystyle{acl_natbib}
\clearpage
\appendix

%

\end{document}